# Vartani Spellcheck – Automatic Context-Sensitive Spelling Correction of OCR-generated Hindi Text Using BERT and Levenshtein Distance

ADITYA PAL[1], ABHIJIT MUSTAFI[2]
*Computer Science and Engineering*
*Birla Institute of Technology, Mesra*
Ranchi, India
[1]aditya.pal.science@gmail.com
[2]abhijit@bitmesra.ac.in

*Abstract*—Traditional Optical Character Recognition (OCR) systems that generate text of highly inflectional Indic languages like Hindi tend to suffer from poor accuracy due to a wide alphabet set, compound characters and difficulty in segmenting characters in a word. Automatic spelling error detection and context-sensitive error correction can be used to improve accuracy by post-processing the text generated by these OCR systems. A majority of previously developed language models for error correction of Hindi spelling have been context-free. In this paper, we present Vartani Spellcheck – a context-sensitive approach for spelling correction of Hindi text using a state-of-the-art transformer – BERT in conjunction with the Levenshtein distance algorithm, popularly known as Edit Distance. We use a lookup dictionary and context-based named entity recognition (NER) for detection of possible spelling errors in the text. Our proposed technique has been tested on a large corpus of text generated by the widely used Tesseract OCR on the Hindi epic Ramayana. With an accuracy of 81%, the results show a significant improvement over some of the previously established context-sensitive error correction mechanisms for Hindi. We also explain how Vartani Spellcheck may be used for on-the-fly autocorrect suggestion during continuous typing in a text editor environment.

## I. Introduction

According to Ethnologue[1], there are more than 7000 languages spoken around the world today. Hindi is ranked third in the list of most spoken languages with an estimated 637 million speakers and is the national language of the country of India. Typed or handwritten text in a language is an important medium for a writer to convey his thoughts. When reading the text, it is necessary to have the correct spelling of words in order to establish meaningful sentences. Ambiguous or meaningless statements affect the flow of reading and may cause the reader to become disinterested.

Due to recent advancements in mobile technology, many readers prefer to use publicly available digitized content for the latest news updates in their preferred native language. More and more reading material is being uploaded on these online digital libraries everyday – be it magazines, newspapers or books. A joint report[2] by Google and KPMG in April 2017 analyzed Indian languages and estimated their internet users. The user base for consuming digital news in Hindi has been predicted to increase 2.5 times – from 55 million in 2016 to 144 million in 2021. The text in these articles is often present as scanned images which cannot be read by a machine directly. Manual extraction of text is time consuming and infeasible. An easier approach is to extract text from these images using Optical Character Recognition (OCR) which is a widely used method to digitize textual content from images or scanned documents. OCR systems for English and other Latin languages have improved significantly over the years and have a very high accuracy for images with reasonable resolution. However, existing OCR systems for Hindi, Bengali, Tamil, Telegu and other highly inflectional Indic languages fail to produce accurate results when compared to English. As pointed out by Vinitha and Jawahar [1], this is mainly due to (a) enormous vocabulary of Indic languages (b) larger number of valid words within a fixed hamming distance (c) lengthier words (d) larger alphabet set with multiple combinations of conjunction between characters, vowels and symbols like halanta, matras etc making it difficult to segment characters in a word (e) lack of resources like named entity recognizers (NERs), Part of Speech Tagging Systems (POS) and other morphological analysers and synthesizers. Since highly accurate OCR systems for Hindi are yet to be designed, post processing the generated OCR text has been regarded as the best option to correct spelling errors.

Most of the traditional error correction methods rely on a straight-forward dictionary lookup approach for error correction of incorrectly spelled words. However, modern Hindi corpora come from multiple sources like chatbot conversations, tweets or Youtube comments which are filled with named entities. Besides, Indic languages like Hindi are morphologically rich and are highly inflectional in nature. Due to these properties, multiple words belonging to separate grammatical categories may be generated from the same root word. Another major inflection point in Hindi is the depiction of gender and seniority of the subject while expressing actions using verbs. This brings us to the importance of "context" in OCR text generated for the Hindi language. Using standard Levenshtein Distance or Longest Common Substring approaches solely for the purpose of error correction causes these systems to overlook contextually correct candidate suggestions, even if the suggested words have correct spelling. Context-sensitive spelling error correction methods are much more efficient for inflectional languages like Hindi. The recent introduction of attention-based transformer models [2] has improved benchmark results of a large number of NLP tasks like Question-Answering and Text Summarization. We decided to solve the spelling error correction problem of OCR generated Hindi text using a Masked Language Model (MLM) approach with BERT [3]. By combining BERT with a Named Entity Recognizer, a lookup dictionary and the Levenshtein Distance algorithm, we designed an automatic context-sensitive spelling error correction mechanism for Hindi language – Vartani Spellcheck. The results from our spellcheck show an improved accuracy. The model can be adapted to autocorrect misspelt words in a fast-typed text-editor environment.

---

[1]https://www.ethnologue.com/
[2]https://assets.kpmg/content/dam/kpmg/in/pdf/2017/04/Indian-languages-Defining-Indias-Internet.pdf

## II. Related Works

This section describes several methods which have been proposed to detect and correct misspelt words in a variety of languages. Hládek et al [4] have surveyed automatic spelling correction techniques and divided all the major approaches into 3 groups – (a) a set of expert rules for error correction (a priori error model), (b) context-based error correction models and (c) learning error patterns from training corpora. The first group of research typically focuses on a set of rules decided by experts and error correction is performed based on only these rules using algorithmic error models like Longest Common Subsequence [5], Levenshtein Distance [6], Damarau-Levenshtein Distance [7] or Phonetic Algorithms only [8]. Context-based error correction models on the other hand derive candidate words by analysing the surrounding word(s) or "context" using components like n-gram models [9], Hidden Markov Models [10] and part-of-speech tagging [11]. The third approach attempts to learn error models from large training corpora using optimization and expectation-maximization algorithms. While learning the word models, context is often taken into account because they involve some of the latest language models like seq2seq Long short-term memory (LSTM) [12], seq2seq Bidirectional LSTM [13] and more recently, transformer models like BERT [14]. This approach has yielded better results in many of the popular languages like English and Chinese.

However, not much effort has been put into implementing these state-of-the-art language models for error correction of low-resource Indic languages like Hindi. A partitioned word dictionary was first proposed by Bansal and Sinha [15]. Another Spell-checker was implemented by Sharma and Jain [16] which uses word-frequency pairs as a language model. HINSPELL was implemented by Singh et al [17] which used Minimum Edit Distance, a weightage algorithm and Statistical Machine Translation. Part-of-Speech tagging has been proposed using rough sets in paper [18]. Recently, "Fuzzy Hindi Wordnet" has been worked on by Jain and Lobiyal [19]. UTTAM, a spelling correction using Supervised Learning has been developed by Jain et al [20]. Singh and Singh have attempted to handle real-word errors using N-gram and Confusion sets [21]. A Deep Learning Approach using character based seq2seq architecture using LSTM has been proposed in SCMIL model by Etoori et al [22]. Srigiri and Saha [23] have approached the problem using Word embeddings – specifically, the Continuous Bag-of-words model. However, transformer-based Hindi spelling error correction has not been worked upon to the best of the authors' knowledge. Hence, we take this opportunity to introduce a new Spellcheck model for Hindi text using BERT.

## III. Proposed Technique

In this section, we discuss the high-level architecture of Vartani Spellcheck as seen in Fig. 1. The input to our spellcheck is the raw text that is generated from an OCR and the output is the text that is obtained after correction. We divide the task into two major steps – error detection and error correction.

### A. Error Detection

The first step of an OCR post-processing model is the detection of spelling errors in the raw OCR output. Error detection has been broadly classified into two categories – (a) Out of Vocabulary (OOV) or "non-word" errors (b) "real-word" errors. OOV words represent invalid words that are not present in the dictionary of the language. For example, the word "heart" when written as "haert" is considered as OOV, whereas "hart" is a valid dictionary word, even though the word may or may not agree in the context in which it is used. When the word "hart" is written in place of "heart", it becomes an example of a real-word error, which is a valid word used out of context. This second category of errors is much harder because traditional dictionary lookup algorithms and non-contextual error correction mechanisms cannot access the nearby words to obtain more contextual information on the subject. In this work, we will be focusing mostly on OOV errors for the Hindi language and formulate a model to detect them. Towards the end of the paper, we will also discuss a possible approach of detecting "real-word" errors using our model which is being worked upon as a future task.

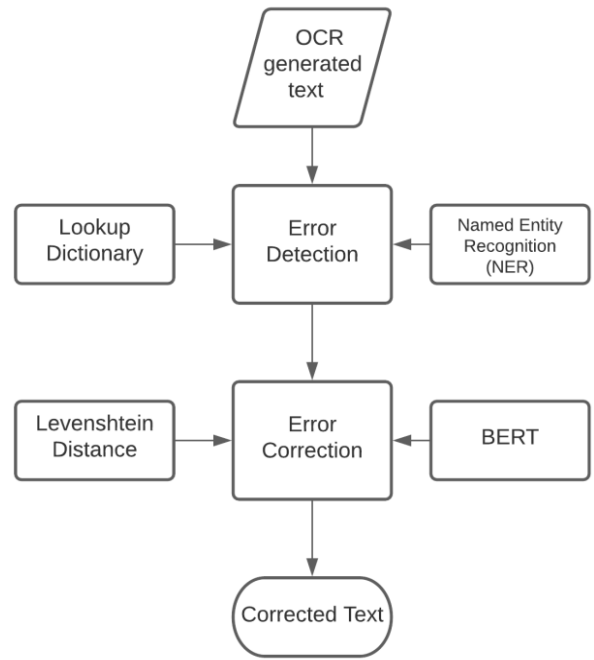

Fig. 1. High-level architecture of Vartani Spellcheck

Error detection in our case is made up of two principal components – a lookup dictionary and an NER.

*1) Lookup Dictionary:* The first component in the error detection setup is a lookup dictionary for Hindi language. For this task, we have gathered a collection of Hindi words from Hindi Hun-Spell dictionary which has 470,000+ unique Hindi words. This dictionary is significantly larger than a Hindi WordNet prepared by the Center For Indian Languages Technology, IIT Bombay which has only about 100,000 unique words.

*2) Named Entity Recognition System (NER):* A named entity recognition module or NER is an NLP component

which is able to identify named or fixed entities in text such as person names, dates, time, locations, quantities, etc. NER systems for English have been worked upon by a number of authors and a few advanced NER systems are available such as spaCy and Stanford NER. Hindi NER systems are yet to catch up with their English counterparts. For our purposes, we have used haptik.ai's open-sourced Chatbot NER [24] as it works quite well for our model. The NER is able to detect the following types of entities:

 *a) Temporal entities* like dates and timestamps

 *b) Numeral entities* like currencies, numerals

 *c) Pattern entities* like email ID, phone number

 *d) Textual entities* like geographical names, cuisines

Using the above two components, we have devised the following criteria for any word to be detected as a misspelt word.

 *a)* Given word should not be a space

 *b)* Word should not be a special character / punctuation

 *c)* Word should not be present in lookup dictionary

 *d)* Word should not be a named entity

Once our model identifies misspelt words, we store those OOV words and their indices with respect to the entire input text in a vector V. We now replace those detected misspelt words with a [MASK] token in the raw text input. This masked text and the vector V now form the input for the next step – the error correction module.

*B. Error Correction*

Error correction is the second step in our Vartani Spellcheck pipeline. Here also, there are primarily two components that are used for correction of the detected errors – a masked language model (MLM), specifically BERT Transformer and Levenshtein Distance.

*1) BERT:* In 2018, a new language represeantation model called BERT (Bidirectional Encoder Representations from Transformers) [3] was introduced by Google AI Language. BERT applies bidirectional training of a popular attention model – Transformer to create a language model. This was different from previous approaches which evaluated text sequences left-to-right only or in combination with right-to-left training. Bidirectional training of the language model was shown to have a better understanding of context. The transformer encoder used in BERT reads the full sequence of words at once, making the model learn all of its surrounding words. The main prediction goal employed in the BERT paper was Masked Language Model or MLM which trains to predict 15% of words in a sequence. 80% of these tokens are replaced with a [MASK] token, 10% with original word and 10% with a random word. MLM tries to predict the original word based on context provided by other unmasked words in sequence. Specifically, given a sentence $x = \{x_i\}$ where i ranges from 1 to the number of words in the sentence L, an MLM models:

$$p(x_j \mid x_1, ..., x_{j-1}, [MASK], x_{j+1}, ..., x_L),$$

where, the token [MASK] is the masked token at the j-th word. Unlike traditional left-to-right language models which calculate $p(x_j \mid x_1, ..., x_{j-1})$, the MLM in BERT uses the full surrounding context to predict the masked word(s) in a sequence of text.

For our model, we have used the original pre-trained masked BERT from the Huggingface Transformers library (bert-base-uncased) [25]. We used the Word Piece tokenizer to represent Hindi words into one or more tokens which are then masked by the MLM. Once the MLM gets trained, we obtain a candidate list of the masked word from the context along with their respective probability values. These candidates represent the list of words which match closely with the surrounding context and are represented by vector $V_c{}^j$ where j denotes the index of the masked token in the sentence. An important hyperparamter to be handled here is the number of words we want to consider in the candidate list. For our dataset, we found that the correct words were predicted with 81% accuracy when 10 candidate words while the percentage dropped to 67% for top 5 words and 55% for top 3 words. Hence we decided to select top 10 words for our candidate list. Once we obtain this candidate list, we move onto the next step of selecting the correct word from the candidate list using Levenshtein Distance.

*2) Levenshtein Distance:* In order to correct spelling errors, we must formulate a method to find the similarity of a given pair of strings. Two widely used algorithms for this purpose are Levenshtein Distance (LD) or Edit Distance and Longest Common Subsequence (LCS). We chose Levenshtein Distance over Longest Common Subsequence because Indic languages like Hindi are highly inflectional in nature. As a result, even the change of just a halant or a vowel can cause a spelling error even though halant and the vowel are very similar optically. For such morphologically rich and inflectional languages Levenshtein Distance is considered as a better approach to calculate similarity of 2 words. LD is used to calculate the minimum number of insertions, deletions or replacements of characters needed to convert an input string S1 into an output string S2. Assume that we have an input string S1 that needs to be transformed into an output string S2 using minimum number of edits known as Minimum Edit Distance (MED). We traverse both the strings from their last character and go in a backward direction up to the first character. Let N be the position in S1 and M be the position in S2 that are currently being looked at. The possible edits and their recursive operations as allowed by LD are as below:

 *a) Insertion* – Add a character at current position
  x = edit_distance(S1, S2, N, M - 1) + 1

 *b) Deletion* – Remove character from current position
  y = edit_distance(S1, S2, N - 1, M) + 1

 *c) Replacement* – Substitute a character with another
  z = edit_distance(S1, S2, N - 1, M - 1)
   + (S1[N - 1] ≠ S2[M - 1])

Minimum Edit Distance MED = minimum (x, y, z)

The operations demonstrated above can be easily calculated for English because every word is perfectly segmented into a fixed set of characters. This is not the case for Devanagari script where a syllable may be written in more than one way. Hence, we make use of a transliteration scheme for Indian languages – WX converter [26]. This scheme is used to represent Devanagari script languages like Hindi in an ASCII format, which helps immensely during Hindi text processing and computation. The scheme uses lowercase English alphabets for short vowels and un-aspirated consonants while aspirated consonants and long vowels are represented by uppercase alphabets.

We now apply Levenshtein Distance on the transliterated values of each of the words from the candidate list of the masked token from the previous step against the incorrect word / token itself. The minimum edit distance of each of these candidate words is calculated and the word having the lowest MED is selected as the replacement word. If two or more words have the same MED, the algorithm automatically picks up the word having the highest probability score.

## IV. WORKING EXAMPLE

In this section, we present an example of how Vartani Spellcheck is able to generate a context-sensitive candidate word list for an incorrect word in a sentence along with their probabilities and then determine the word having the least MED among those words. Let us consider a statement from the Ramayana (Hindi Epic) – "राम ने खाना खाया" which correctly translates to "Ram ate food". An OCR had incorrectly generated the text "राम ने खाना राया" where the third word in the sentence is an OOV word with a spelling error. Our spelling detection module accurately identifies this mistake and does not assign it as a named entity. Now our BERT model generates a list of top 10 candidate words. Here, we show only the top 5 recommendations in Fig. 2 as the rest have very low probabilities and higher MEDs. As seen from the example in the table, Vartani Spellcheck was able to accurately correct the misspelt word by using contextual information and Edit Distance on the transliterated values of the candidate words.

TABLE I. CANDIDATE WORDS OF VARTANI SPELLCHECK

| Candidate Word | Probability | Transliteration | MED from rAyA |
|---|---|---|---|
| खाया | 0.4191 | KAyA | 1 |
| बनाया | 0.2359 | banAyA | 3 |
| खिलाया | 0.1257 | KilAyA | 3 |
| लाया | 0.0124 | lAyA | 1 |
| पकाया | 0.0113 | pakAyA | 3 |

Fig. 2. List of candidate words, their probabilty as assigned by BERT, their transliterations and Minimum Edit Distance with respect to the OOV word "राया" which transliterates to rAyA

## V. RESULTS

In this section, we present the experimental setup and the results obtained after running Vartani Spellcheck on a self-generated dataset. To create the dataset, we have taken a low-resolution online scanned version of the Hindi epic "Ramayana". After obtaining the copy, we extracted the first 40 pages of the scanned copy and ran the popular Tesseract OCR on those pages. The resultant text was split into individual sentences and manually searched for errors in optical character recognition. Out of 2236 sentences, we found 939 sentences having at least a single word error. A manual correction of these errors was done and used for comparison against the output generated by Vartani spellcheck. We processed each of these incorrect 939 sentences using Vartani Spellcheck and the results are summarized in Fig. 3. From the table, we can see that our model performs quite well when we consider a higher number of words in candidate list. For 10 candidate words, we see that 81% of these OOV spelling errors are corrected accurately. However, this same percentage drops to 67% when top 5 words are considered in the candidate set and there is a further drop to 55% for top 3 words. Nevertheless, these results are quite promising as we do not necessarily need to compromise the number of candidate words for the accuracy obtained. So, the experiments suggest that 81% of OCR-generated OOV or non-word errors can be accurately corrected by our Vartani Spellcheck. This is a slight improvement over the 71.67% accuracy obtained by the Continuous-Bag-Of-Word approach proposed by Srigiri and Saha [23], although they have taken into account only top 3 words in their candidate list.

TABLE II. ACCURACY OF VARTANI SPELLCHECK

| S.No. | Number of candidate words | Accuracy in % |
|---|---|---|
| 1 | 1 | 28.70 |
| 2 | 3 | 55.46 |
| 3 | 5 | 66.97 |
| 4 | 10 | 81.03 |
| 5 | 20 | 85.78 |

Fig. 3. Accuracy of Vartani Spellcheck based on first 40 pages of Tesseract OCR generated Hindi text of the epic Ramayana

## VI. CONCLUSION AND FUTURE WORK

In this paper, we present a new approach to tackle the spelling error correction problem of a low-resource Indic language using state-of-the-art transformer model – BERT. The results obtained are promising and may be improved by fine-tuning the BERT model even further. We are yet to explore the power of newer and advanced transformer models like ALBERT and RoBERTa. We are also planning to fine-tune Indic-BERT [27] for our spelling correction task since it is pretrained on a large corpus of Hindi text and may yield better candidates for the masked tokens. Although we have proposed a model for the most common Indic language and the national language of India, we would like to extend Vartani Spellcheck to other low-resource languages like Bengali, Tamil, Telugu, Marathi, Gujarati and many others. For the moment, we have only focused on Out of Vocabulary (OOV) words for spelling error detection. Work is in progress to fine-tune a transformer model for detection of real-word errors. Although it is an uphill task with very little resources, we have already created a dataset for classification of grammatically correct and incorrect sentences from a text corpus. We hope to incorporate as many Indic languages as possible in the next iteration of Vartani Spellcheck with real-word errors handled automatically. Lastly, we are attempting to work on a character level fine-tuning of BERT which may produce even better results for our given task. Vartani Spellcheck can also be used for "on-the-fly" spelling error correction in a text editor environment. For this, we would need to look ahead up to a few words from the incorrect word or the entire statement before we run our error correction model. The misspelt OOV word may be highlighted at a real-time pace, thus enabling users to make a choice to manually correct themselves or wait for a few words until autocorrect kicks in. In a fast-typed editor environment, a user is likely to type multiple words ahead of the incorrect word, making Vartani Spellcheck feasible.

ACKNOWLEDGMENT

We are extremely thankful to Google Colaboratory [24] and their powerful hardware which was used to run our BERT models. We would also like to thank the creators of publicly available Hindi datasets which were used extensively in our research.

REFERENCES

[1] V. S. Vinitha and C. V. Jawahar, "Error Detection in Indic OCRs," 2016 12th IAPR Workshop on Document Analysis Systems (DAS), Santorini, 2016, pp. 180-185, doi: 10.1109/DAS.2016.31.

[2] Vaswani, Ashish, et al. "Attention is all you need." *Advances in neural information processing systems* 30 (2017): 5998-6008.

[3] Devlin, J., Chang, M. W., Lee, K., & Toutanova, K. (2018). Bert: Pre-training of deep bidirectional transformers for language understanding. *arXiv preprint arXiv:1810.04805*.

[4] Hládek, Daniel, Ján Staš, and Matúš Pleva. "Survey of Automatic Spelling Correction." *Electronics* 9.10 (2020): 1670.

[5] Taghva, Kazem, and Eric Stofsky. "OCRSpell: an interactive spelling correction system for OCR errors in text." *International Journal on Document Analysis and Recognition* 3.3 (2001): 125-137.

[6] Van Delden, Sebastian, David Bracewell, and Fernando Gomez. "Supervised and unsupervised automatic spelling correction algorithms." *Proceedings of the 2004 IEEE International Conference on Information Reuse and Integration, 2004. IRI 2004.*. IEEE, 2004.

[7] Hagen, M., Potthast, M., Gohsen, M., Rathgeber, A., & Stein, B. (2017, August). A large-scale query spelling correction corpus. In *Proceedings of the 40th International ACM SIGIR Conference on Research and Development in Information Retrieval* (pp. 1261-1264).

[8] Kondrak, Grzegorz, and Tarek Sherif. "Evaluation of several phonetic similarity algorithms on the task of cognate identification." *Proceedings of the Workshop on Linguistic Distances*. 2006.

[9] Ahmed, Farag, Ernesto William De Luca, and Andreas Nürnberger. "Revised n-gram based automatic spelling correction tool to improve retrieval effectiveness." *Polibits* 40 (2009): 39-48.

[10] Stüker, Sebastian, Johanna Fay, and Kay Berkling. "Towards Context-Dependent Phonetic Spelling Error Correction in Children's Freely Composed Text for Diagnostic and Pedagogical Purposes." *Twelfth annual conference of the international speech communication association*. 2011.

[11] Vilares, J., Alonso, M. A., Doval, Y., & Vilares, M. (2016). Studying the effect and treatment of misspelled queries in Cross-Language Information Retrieval. *Information Processing & Management*, 52(4), 646-657.

[12] Evershed, John, and Kent Fitch. "Correcting noisy OCR: Context beats confusion." *Proceedings of the First International Conference on Digital Access to Textual Cultural Heritage*. 2014.

[13] Zhou, Yingbo, Utkarsh Porwal, and Roberto Konow. "Spelling correction as a foreign language." *arXiv preprint arXiv:1705.07371* (2017).

[14] Sun, Yifu, and Haoming Jiang. "Contextual Text Denoising with Masked Language Models." *arXiv preprint arXiv:1910.14080* (2019).

[15] Bansal, Veena, and R. M. K. Sinha. "Partitioning and searching dictionary for correction of optically read Devanagari character strings." *International journal on document analysis and recognition* 4.4 (2002): 269-280.

[16] Amit Sharma and Pulkit Jain. "Hindi spell checker." *Indian Institute of Technology Kanpur* (2013).

[17] Kaur, B., and H. Singh. "Design and implementation of HINSPELL—Hindi spell checker using hybrid approach." *Int. J. Sci. Res. Manage* 3.2 (2015): 2058-2061.

[18] Gupta, J. P., Devendra K. Tayal, and Arti Gupta. "A TENGRAM method based part-of-speech tagging of multi-category words in Hindi language." *Expert Systems with Applications* 38.12 (2011): 15084-15093.

[19] Jain, Amita, and D. K. Lobiyal. "Fuzzy Hindi WordNet and word sense disambiguation using fuzzy graph connectivity measures." *ACM Transactions on Asian and Low-Resource Language Information Processing (TALLIP)* 15.2 (2015): 1-31.

[20] Jain, Amita, et al. ""UTTAM" An Efficient Spelling Correction System for Hindi Language Based on Supervised Learning." *ACM Transactions on Asian and Low-Resource Language Information Processing (TALLIP)* 18.1 (2018): 1-26.

[21] Singh, Shashank, and Shailendra Singh. "Handling Real-Word Errors of Hindi Language using N-gram and Confusion Set." *2019 Amity International Conference on Artificial Intelligence (AICAI)*. IEEE, 2019.

[22] Etoori, Pravallika, Manoj Chinnakotla, and Radhika Mamidi. "Automatic spelling correction for resource-scarce languages using deep learning." *Proceedings of ACL 2018, Student Research Workshop*. 2018.

[23] Srigiri, Shashank, and Sujan Kumar Saha. "Spelling Correction of OCR-Generated Hindi Text Using Word Embedding and Levenshtein Distance." *International Conference on Nanoelectronics, Circuits and Communication Systems*. Springer, Singapore, 2018.

[24] https://www.haptik.ai/tech/open-sourcing-chatbot-ner/

[25] Wolf, Thomas, et al. "Transformers: State-of-the-art natural language processing." *Proceedings of the 2020 Conference on Empirical Methods in Natural Language Processing: System Demonstrations*. 2020.

[26] https://github.com/irshadbhat/indic-wx-converter

[27] Kakwani, Divyanshu, et al. "iNLPSuite: Monolingual Corpora, Evaluation Benchmarks and Pre-trained Multilingual Language Models for Indian Languages." *Proceedings of the 2020 Conference on Empirical Methods in Natural Language Processing: Findings*. 2020.

[28] Bisong, Ekaba. "Google Colaboratory." *Building Machine Learning and Deep Learning Models on Google Cloud Platform*. Apress, Berkeley, CA, 2019. 59-64.